\documentclass[10pt,twocolumn,letterpaper,table]{article}

\usepackage{iccv}
\usepackage{xcolor}
\usepackage{times}
\usepackage{epsfig}
\usepackage{graphicx}
\usepackage{amsmath}
\usepackage{amssymb}
\usepackage{booktabs}
\usepackage{multirow}
\usepackage{subcaption}

\usepackage{hyperref}
\usepackage{url}
\usepackage{multirow}
\usepackage{tikz}
\usepackage{comment}
\usepackage{color}
\usepackage{cuted}
\usepackage{capt-of}
\usepackage{comment}
\usepackage{breqn}
\usepackage{makecell}
\usepackage{enumitem}

\iccvfinalcopy % *** Uncomment this line for the final submission

\def\iccvPaperID{****} % *** Enter the ICCV Paper ID here

\ificcvfinal\fi
\newif\ifdraft
\draftfalse
\ifdraft
    \newcommand{\rg}[1]{{\color{red}{[rahul: #1]}}}
    \newcommand{\jmj}[1]{{\color{green}{[jose: #1]}}}
    \newcommand{\al}[1]{{\color{blue}{[andreas: #1]}}}
 \else
    \newcommand{\rg}[1]{}
    \newcommand{\jmj}[1]{}
    \newcommand{\al}[1]{}
\fi

\begin{document}

%%%%%%%%% TITLE
\title{ReBotNet: Fast Real-time Video Enhancement}

\author{Jeya Maria Jose Valanarasu\textsuperscript{1,3$\ast$}~~~~~~Rahul Garg$^{1}$~~~~~~Andeep Toor$^{1}$~~~~~~Xin Tong$^{1}$~~~~~~Weijuan Xi$^{1}$\\
Andreas Lugmayr$^{2}$~~~~~~Vishal M. Patel$^{3}$~~~~~~Anne Menini$^{1}$ \\
$^1$Google~~~~~~$^2$ETH Zurich~~~~~~$^3$Johns Hopkins University\\
}

\maketitle
\def\thefootnote{*}\footnotetext{Parts of the work was done during an internship at Google.}\def\thefootnote{\arabic{footnote}}
\ificcvfinal\thispagestyle{empty}\fi

%%%%%%%%% ABSTRACT
\begin{abstract}
   %Real-time video enhancement improves the quality of live video streams and video calls providing users a pleasant experience regardless of their network connection or webcam quality. 
   Most video restoration networks are slow, have high computational load, and can't be used  for real-time  video enhancement. In this work, we design an efficient and fast framework to perform real-time video enhancement for practical use-cases like live video calls and video streams. Our proposed method, called \textbf{Re}current \textbf{Bot}tleneck Mixer \textbf{Net}work (\textbf{ReBotNet}), employs a dual-branch framework. The first branch learns spatio-temporal features by tokenizing the input frames along the spatial and temporal dimensions using a ConvNext-based encoder and  processing these abstract tokens using a bottleneck mixer. To further improve temporal consistency, the second branch employs a mixer directly on tokens extracted from individual frames. A common decoder then merges the features form the two branches to predict the enhanced frame. In addition, we propose a recurrent training approach where the last frame's prediction is leveraged to efficiently enhance the current frame while improving temporal consistency.  To evaluate our method, we curate two new datasets that emulate real-world video call and streaming scenarios, and show extensive results on multiple datasets where ReBotNet outperforms existing approaches with lower computations, reduced memory requirements, and faster inference time. Project site: \href{https://jeya-maria-jose.github.io/rebotnet-web/}{https://jeya-maria-jose.github.io/rebotnet-web/}
   
\end{abstract}

%%%%%%%%% BODY TEXT
\section{Introduction}

\begin{figure}
    \centering
    \includegraphics[width=1\linewidth]{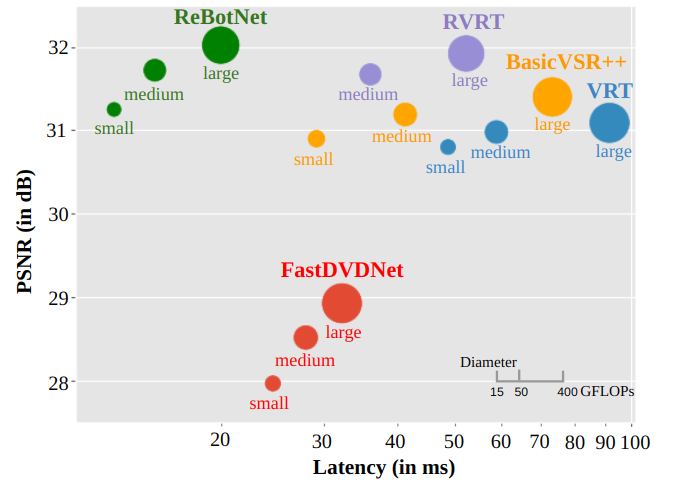}
    \vspace{-2 em}
    \caption{A comparison between the performance of ReBotNet with state-of-the art video restoration networks across different FLOPs regimes on a NVIDIA A100 GPU  for PortraitVideo dataset. ReBotNet is observed to give the best performance while having the least latency.}
    \al{Is this the average over both datasets?}
    \label{fig:chart1}
\vspace{-1.5 em}
\end{figure}

Video enhancement has several use-cases in surveillance \cite{shen2022image, ding2020deep, rajan2016enhancement}, cinematography \cite{wan2022bringing, iizuka2019deepremaster}, medical imaging \cite{katsaros2022multi,stetson1997lesion}, virtual reality \cite{wang2022virtual, pearl2022nan, han2007real, vassallo2018augmented}, sports streaming \cite{chang2001real,zhao2022motion}, and video streaming \cite{zhang2020improving, zhang2020improving}. It also facilitates downstream tasks such as analysis and interpretation \cite{rao2012survey}, e.g., it improves accuracy of facial recognition algorithms, allows doctors to diagnose medical conditions more accurately, and helps in better sports analysis by understanding player movements and tactics.  Also, the recent rise of hybrid work has led to an immense increase in video conferencing, where poor video quality due to a low quality camera, poor lighting conditions, or a bad network connection can obscure non-verbal cues and hinder communication and increase fatigue \cite{doring2022videoconference}.  %Video conferencing saves time and effort and has been proven to increase productivity. 
%In these scenarios, how well a person is represented is a crucial aspect for their participation and effectiveness in the meeting. It is common for video calls to be hindered by low quality of camera, transmission noise, or poor lighting conditions. 
Thus, there lies a significant interest in developing methods that can perform real-time video enhancement. 

%Video restoration refers to the process of improving the quality of a damaged video usually degraded by a single degradation like noise, camera quality etc. %In restoration, the goal is to invert a known degradation operations applied to videos. 
Unlike individual restoration tasks like denoising \cite{elad2023image, tian2019deep}, deblurring \cite{zhang2022deep, sahu2019blind}, super-resolution \cite{wang2020deep, liu2022video} which focus on restoring videos affected by a single degradation;  generic video enhancement techniques focus on improving the overall quality of videos and make them look better \cite{xue2019video}. 
In this setup, there are multiple degradations that can interact in a complex way, e.g., compression of a noisy video, camera noise, motion blur etc. mirroring the real world scenarios. %can be unknown or even a combination of multiple unknown degradations. In real-world, there is a high chance we come across videos degraded by multiple degradations. %Most of the previous works focus on video restoration and not video enhancement.
%There are several video restoration methods in literature developed which look into specific problems like deblurring, super-resolution, and denoising. 
Video restoration methods can be adopted for video enhancement by training on a dataset that includes multiple degradations.
However, from our experiments we found that they are  computationally complex and have a high inference time and are not suitable for real-time applications.
%For real-time applications like live video calls and streams, the latency of the model plays a pivotal role as a slow model results in low frames per rate which affects the video feed.
Also, many methods take past and future frames as input which will introduce latency in streaming video.
%take a set of frames as input and predict the center frame. This is not possible in live scenarios like video calls where we do not have access to the future frame. \rg{Do some of them also take future frames as input? We should emphasize that we can't do so in streaming case.}\jmj{ That is a great point, added.}

In this paper, we develop an efficient video enhancement network that achieves state of the art results and enables real-time processing. At the core of our method is a novel architecture using convolutional blocks at early layers and MLP-based blocks at the bottleneck. Following \cite{srinivas2021bottleneck} which uses a convolutional encoder for initial feature extraction followed by a transformer network in the bottleneck, we propose a network where the initial layers extract features using ConvNext \cite{liu2022convnet}  blocks and a bottleneck consisting of MLP mixing blocks \cite{tolstikhin2021mlp}. This design avoids quadratic computational complexity of vanilla attention \cite{vaswani2017attention}, while maintaining a good performance. We also tokenize the input frames in two different ways to enable the network to learn both spatial and temporal features. Both these token sets are passed through separate mixer layers to learn dependencies between these tokens. We then use a simple decoder to predict the enhanced frame. %This setup of using a convnext for feature extraction and a bottleneck mixer for processing and aggregate all the information helps us to attain a good performance with a smaller compute.
%Second, we propose a technique to further improve the performance while fixing the network architecture so that we could maintain the computational complexity. 
To further improve efficiency and improve temporal consistency, we exploit the fact that real world videos typically have temporal redundancy implying that the prediction from previous frame can help current frame's prediction.
To leverage this redundancy, we use a frame-recurrent training setup where the previous prediction is used as an additional input to the network. This helps us carry forward information to the future frames while being more efficient than methods that take a stack of multiple frames as input. We train our proposed network in this recurrent way and term our overall method \textbf{Re}current \textbf{Bot}tleneck Mixer \textbf{Net}work (ReBotNet). 

 To evaluate our method, we curate and introduce two new datasets for video enhancement. The existing video restoration datasets focus on a single task at a time, e.g., denoising (DAVIS \cite{khoreva2019video}, Set8 \cite{tassano2019dvdnet}, etc.), deblurring (DVD \cite{su2017deep}, GoPro \cite{nah2017deep}, etc.), and super-resolution (REDS \cite{nah2019ntire}, Vid4 \cite{liu2013bayesian}, Vimeo-90k-T \cite{xue2019video}, etc.). These datasets do not emulate the real-world case where the video is degraded by a mixture of many artifacts. Also, rise in popularity of video conferencing calls for datasets that have semantic content similar to a typical video call. Single image enhancement methods are often studied on face images~\cite{liu2018large, karras2019style} because human perception is very sensitive to even slight changes in faces. However, a comparable dataset for video enhancement research has yet to be established.  To this end, we curate a new dataset called \textit{PortraitVideo} that contains cropped talking heads of people and their corresponding degraded version obtained by applying multiple synthetic degradations. The second dataset, called \textit{FullVideo}, contains a set of degraded videos without face alignment and cropping.
 %Note that these two video enhancement datasets are different from previous video restoration as they are carefully curated with a combination of degradations occuring at practical scenarios like video calls and streams. 
 We conduct extensive experiments on these datasets and show that we obtain better performance with less compute and faster inference than recent video restoration frameworks. In particular, our method is 2.5x faster while either matching or in some cases obtaining a PSNR improvement of 0.2 dB over previous SOTA method. This shows the effectiveness of our proposed approach and opens up exciting possibilities of deploying them in real-time applications like video conferencing. 
 %\rg{Overall, the motivation for new datasets seems a bit weak. It seems we are claiming that we need them for two reasons 1) multiple degradations 2) Containing faces, similar to videos calls. For 1) I think the reviewers will still want to see single degradation vs multiple degradation comparison. Do we not do so well if there's a single degradation? If so, why? Is the problem too easy? For 2), I was wondering if we can claim that for an ill-posed problem like video restoration, a restricted domain, e.g., face videos, allows learning based methods to learn domain specific priors. Though we need to be careful so that the reviewers don't ask us to compare with face restoration methods.} \jmj{We can still show results on single degradation but I feel it is beyond the scope of this paper. Going in that direction would require us to show SOTA on a lot of datasets which might not be that straight-forward. Face-specific prior will again open up a lot of questions and suggestions like we can include face priors like from GFP-GAN.}

 \al{Add reason for new datasets. e.g. Advancements in video enhancement have resulted in outdated datasets. Despite some datasets being of high resolution, the ground truth images frequently contain artifacts such as blur. To overcome this challenge, we have developed and published two new datasets with superior quality. First, the FullVideo contains diverse videos of high quality. Single image enhancement methods are often studied on face images~\cite{celeba,ffhq} because human perception is very sensitive to even slight changes in faces. However, a comparable dataset for video enhancement research has yet to be established. To address this gap, we propose PortraitVideo.}\jmj{added}

In summary, we make the following major contributions:

\begin{itemize}[topsep=0pt,noitemsep,leftmargin=*]

	\item We work towards \textbf{real-time} video enhancement, with a specific focus on practical applications like video calls and live streaming. 
	
	\item We propose a new method: Recurrent Bottleneck Mixer Network (ReBotNet) , an efficient deep neural network architecture for real-time video enhancement.
	
	\item We curate two new video enhancement datasets: \textit{PortraitVideo}, \textit{FullVideo} which emulate practical video enhancement scenarios. \al{If we publish the data, it would strengthen this contribution if we state that. The strongest case would be to provide an anonymous link to the data (e.g. using https://archive.org/create/ or github release files)}
 
 \item We perform extensive experiments where we find that ReBotNet matches or exceeds the performance of baseline methods while being significantly faster.
	
\end{itemize}

% \section{Related Works}

% \subsection{Video Restoration}
% Over the years, a number of methods have been proposed to address these issues and restore the quality of the video. Similar to image restoration [], learning-based methods, especially ConvNet-based methods, are the primary workhorse for video restoration. sliding window-based and recurrent methods. Sliding window-based methods often takes a short sequence of frames as input and merely predict the center frame. Although some works [43] predict multiple frames, they still focus on the reconstruction of the center frame during training and testing.

% One popular approach for video restoration is based on convolutional neural networks (CNNs), which have been shown to be effective at modeling the spatial relationships between pixels in an image. Many existing CNN-based methods use a supervised learning approach, where the network is trained on a dataset of degraded and reference videos to learn a mapping from the degraded video to the reference video. However, these methods typically do not consider the temporal dependencies between video frames, which can be crucial for restoring the quality of the video.

\section{Related Work}

Image and video restoration \cite{dong2014learning, fan2017balanced, fan2020scale, guo2020closed, zhang2018learning, zhang2019residual, zhou2019spatio, yasarla2020exploring, perera2022transformer, yang2021ntire, yi2021omniscient, yi2019progressive} is a widely studied topic where CNN-based methods have been dominating over the past few years. For video restoration, most CNN-based methods take a sliding window approach where a sequence of frames are taken as input and the center frame is predicted \cite{su2017deep, tassano2019dvdnet, tassano2020fastdvdnet, tian2020tdan, wang2019edvr, zhou2019spatio}. To address motion between frames, many methods explicitly focus on temporal alignment \cite{chan2021basicvsr,zhu2022deep,chan2021understanding, tian2020tdan, wang2019edvr}, with optical flow being a popular alignment method \cite{caballero2017real,kappeler2016video, liu2017robust, liu2017robust, tao2017detail,xue2019video}.
%SpyNet \cite{ranjan2017optical} is used as the pre-trained optical flow estimation model that is used to align various video restoration tasks \cite{xue2019video}. 
Dynamic upsampling filters \cite{jo2018deep}, spatio-temporal transformer networks \cite{kim2018spatio}, and deformable convolution \cite{tian2020tdan} have been proposed for multi-frame optical flow estimation and warping. Aside from sliding window approaches, another widely used technique is a recurrent framework where bidirectional convolutional neural networks warp the previous
frame prediction onto the current frame \cite{chan2021basicvsr, chan2022basicvsr++, fuoli2019efficient, haris2019recurrent, huang2015bidirectional, isobe2020video, nah2019recurrent, park2020bmbc, son2021recurrent, zhong2020efficient,sajjadi2018frame}. These recurrent methods usually use optical flows to warp the nearby frames to create the recurrent mechanism. Unlike these works that require compute intensive optical flow, we develop a simple and efficient frame-recurrent setup with low computational overhead.  As most of these methods use synthetic datasets, recent works have looked into adopting these methods for real-world application \cite{yang2021real, chan2022investigating}\rg{Do these still focus on one degradation? If so, mention that}. One recent work attempted to solve multiple degradation problem that includes  blur, aliasing and low resolution with one model \cite{cao2022towards} but it is still computationally intensive.

Since the introduction of transformers \cite{vaswani2017attention} for visual recognition \cite{dosovitskiy2020image, liu2021swin}, transformers have been widely adopted for many restoration tasks \cite{valanarasu2022transweather, chen2021pre, wang2021uformer, liang2021swinir, tan2021sdnet, qin2021etdnet, zhao2021hybrid, cao2022ciaosr, zhong2022blur, shi2022rethinking}. Deformable attention \cite{wang2023stdan} has been proposed for video super-resolution. Video restoration transformer (VRT) introduced a parallel frame prediction model leveraging long-range temporal dependency modelling abilities of transformers \cite{liang2022vrt}. Recurrent video restoration transformer (RVRT) \cite{liang2022recurrent} introduced a globally recurrent framework with processing neighboring
frames. At the time of writing, it is worth mentioning that RVRT stands as the SOTA method for most video restoration datasets. Unlike above methods, we focus on developing real-time solutions for generic video enhancement with a focus on practical applications like live video calls.

\section{Method}
\begin{figure*}
    \centering
    \includegraphics[width=0.9\linewidth, page=1]{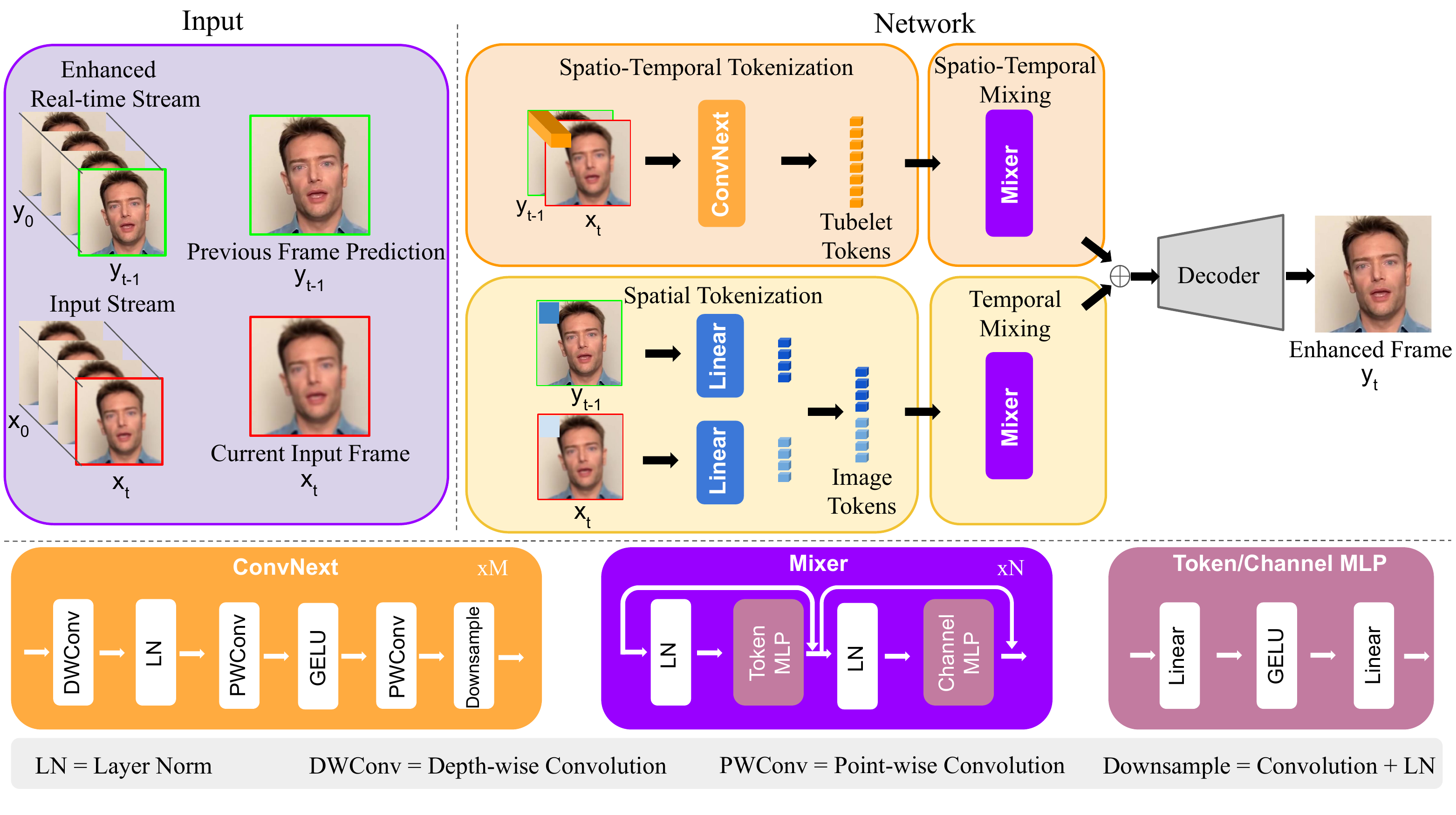}
    \vspace{-1 em}
    \caption{Overview of the proposed Recurrent Bottleneck Mixer Network. The inputs to the network are the previous frame prediction and the current input frame. These are tokenized in two different ways: Tubelet tokens and image tokens . The tubelet tokens are processed using a Mixer to learn spatio-temporal features while image tokens are processed using a Mixer to learn temporal features. These features are passed through an upsampling decoder to get the output enhanced frame.}
    \label{fig:overview}
\end{figure*}

\subsection{Recurrent Bottleneck Mixer}
 Transformers \cite{dosovitskiy2020image} form the backbone of current state of the art video restoration methods \cite{liang2022vrt, liang2022recurrent} due to their ability to model long-range dependencies but suffer from high computational cost due to the quadratic complexity of attention mechanism. Attention with linear complexity \cite{wang2020linformer,zhang2022patchformer, katharopoulos2020transformers} reduces performance while still not achieving real-time inference.
 %There have been works which work towards linearizing the attention mechanism \cite{wang2020linformer,zhang2022patchformer, katharopoulos2020transformers}. \rg{Why are they bad then? Drop in performance unlike MLP-mixers?}\jmj{yeah, they are still not efficient} 
 On the other hand, \cite{zhai2021attention, liu2021pay, wang2022shift} show that attention can be replaced by other mechanisms with marginal regression in quality, e.g., \cite{tolstikhin2021mlp}  replaces self-attention with much more efficient token mixing multi-layer perceptrons (MLP-Mixers).
 %to avoid the quadratic complexity of transformers.
 %They achieve a better throughput without much drop in performance.  %\cite{liu2022convnet} which work towards reducing the computation of transformers while trying to main performance. 
 %Mixers replace the self-attention with token mixing multi-layer perceptrons (MLP) to avoid the quadratic complexity of transformers. 
 %Adopting MLPs for self-attention in transformers 
 Mixers have been subsequently shown to be useful for multiple tasks  \cite{touvron2022resmlp, valanarasu2022unext, yu2022s2, qiu2022mlp, tu2022maxim, ma2022rethinking}. 
  %In our efforts to reduce the latency of video enhancement networks, we initially attempted to use MLP-Mixers directly for video enhancement.
  However, Mixers do not work out-of-the box for video enhancement, as (i) they lead to a significant regression in quality (in our experiments in supplementary material) compared to transformer-based approaches, and (ii) while more efficient, they still do not yield real-time inference on high resolution imagery.
  %In our experiments, we observed that MLP-Mixers tend to exhibit a noticeable decline in quality for video enhancement compared to transformer-based approaches. \al{Is this decine shown in the ablation study?} Using Mixers directly on large size images still takes a lot of compute and makes it difficult to achieve real-time speed. 
  %Hence, we develop a new efficient backbone for video enhancement that leverages MLP-mixers at its core.
  Also, videos are processed using transformers by either representing them as tubelets or patch tokens \cite{arnab2021vivit}. However, tubelet tokens \cite{arnab2021vivit} and image tokens \cite{dosovitskiy2020image} can be complementary with different advantages and disadvantages. Tubelet tokens can compactly represent spatio-temporal patterns. On the other hand, image tokens or patch tokens extracted from an individual frame represents only spatial features without spending capacity on modeling motion cues. These issues motivates us in developing a new backbone for video enhancement with mixers at its core while combining tubelets and image tokens in a single efficient architecture.
%ConvNext, on the other hand sticks to a convolutional neural network (ConvNet) based architecture but employs many tricks from the development of transformers. Both these approaches prove to be useful and achieve a comparable performance on image recognition tasks with lesser computation than vision transformers (ViT).

  After motivating the design, we now explain our proposed network architecture: Recurrent Bottleneck Mixer Network (ReBotNet) in detail. First, we give an idea about the overall network architecture and then delve into the details of tokenization, bottleneck, and the recurrent setup.  An overview of ReBotNet can be found in Fig.~\ref{fig:overview}. ReBotNet takes two inputs: the previous predicted frame ($y_{t-1}$) and the current frame ($x_{t}$). 
  %Please note that this is not a strict criteria as we can easily extend ReBotNet for more than 2 frames. %We choose this simple 2 frame approach for most of experiments as we focus on reducing the complexity. 
  We use an encoder-decoder architecture where the encoder has two branches. The first 
 branch focuses on  spatio-temporal mixing where we tokenize the input frames as tubelets and then process these spatio-temporal features using mixers in the bottleneck. 
 The output features of this mixer block has information processed along both the spatial and temporal dimensions.
 The second branch extracts just the spatial features using linear layers from individual frames. These tokens contain only spatial information as the frames are processed independently. These spatial features are forwarded to another mixer bottleneck block which learns the inter-dependencies between these tokens. This mixer block captures temporal information by extracting the relationship between tokens from individual frames, thereby encoding the temporal dynamics.
 The resultant features from both branches are added and are forwarded to a decoder which consists of transposed convolutional layers to upsample the feature maps to the same size as of the input. We output a single prediction image  ($y_{t}$) which is the enhanced image of the current frame  ($x_{t}$).

\subsection{Encoder and Tokenization}

Tokenization is an important step in pre-processing data for transformer-based methods as it allows the model to work with the input data in a format that it can understand and process \cite{qian2022makes}. For our network, we use two different ways of doing tokenization: i) tubelet tokens and ii) image tokens.

\noindent \textbf{Branch 1 - Tubelet tokens:} Tubelet tokens are extracted across multiple frames, in our case, the current frame and the previous predicted frame, and 
encode spatio-temporal data. Convolutional layers can be advantageous in extracting tokens as they can capture more informative features compared to linear layers due to their inductive bias 
 \cite{xie2021segformer}.
%for feature extraction can result in more efficient encoding of tokens as it allows for better extraction of relevant features. 
Hence, we stack the input images: $y_{t-1}$,  $x_{t}$ across the channel dimension and directly forward them to ConvNext blocks \cite{liu2022convnet}, which are more efficient and powerful than vanilla convolutional layers.
%The convolution operation in this case is applied across both the spatial and temporal dimensions, and the input is processed as tubelets. As a result, the extracted tokens contain information from both the spatial and temporal domains.
%We choose ConvNext to extract the features as they are more efficient and powerful than basic convolutional blocks. 
Each ConvNext block consists of a depth-wise convolution layer \cite{chollet2017xception} with kernel size of $7 \times 7$, stride $1$ and padding $3$ followed by a layer normalization \cite{ba2016layer} and a point-wise convolution function.  The output of this is activated using GeLU \cite{hendrycks2016gaussian} activation and then forwarded to another point-wise convolution to get the output. More details of this why this exact setup is followed can be found in supplementary. We also have downsampling blocks after each level in the ConvNext encoder. These tubelet tokens compromise the first branch of ReBotNet where we do spatio-temporal mixing. These tokens are further processed using a bottleneck mixer to enhance the  features and encode more spatio-temporal information.

\noindent \textbf{Branch 2 - Image tokens:} The individual frames $y_{t-1}$,  $x_{t}$ are from different time steps. Although tubelet tokens encode temporal information, learning additional temporal features can only improve the stability of the enhanced video and help get clearer details for enhancement. We do this by extracting individual image tokens and learn the correspondence between them. To this end, we tokenize the images individually by converting them into patches and using linear layers like in ViT \cite{dosovitskiy2020image}. In this branch, we use linear layers instead of ConvNext blocks for the sake of efficiency although ConvNext blocks extract more representative and useful features. 
%and in order to learn compleTo simplify the model and prioritize maintaining temporal consistency, we use linear layers instead of ConvNext blocks.
The main goal of this block is to ensure that the temporal features of the input data remain consistent.  Note that high quality spatial features necessary for enhancing spatial quality, is handled in the first branch. To this end, the mixer bottleneck learns to encode the temporal information between these image tokens extracted from individual frames. \rg{I thought the second branch learns spatial features, e.g., see abstract?}\jmj{yes, it extracts spatial features but that facilitates temporal mixing.}
%\rg{computational complexity? I thought (fully connected?) linear layers are more compute intensive?}\jmj{I think it depends on how much tokens we extract and the conv kernel size and stride. also here we have conv blocks in first branch; so just one linear layer is less complex compared to that. Also, check table 3.}

We ensure that the tubelet tokens and image tokens have the same dimensions of $N \times C$, where $N$ is the number of tokens and $C$ is the number of channel embeddings. To achieve this, we max-pool image tokens to match the dimensions of tubelet tokens.
%using max-pooling to match their $N$ value with that of the tubelet tokens and use the same value of $C$ for both types of tokens.

% The encoder consists of multiple ConvNext blocks. We pick ConvNext blocks over basic ConvNet blocks as it has been shown that they are both efficient and effective than ConvNets. Each ConvNext block first consists of a depth-wise convolution block with kernel size of $7 \times 7$, stride $1$ and padding $3$. Using a large kernel size is to have a larger receptive field similar to non-local attention. It was observed in \cite{liu2022convnet} that
% the benefit of larger kernel sizes reaches a saturation point at at $7 \times 7$. It is followed by a layer normalization and a point-wise convolution function. The point-wise convolution is basically a convolution layer with kernel size $1 \times 1$. The output of this is activated using GeLU activation and then forwarded to another point-wise convolution to get the output. More details of this why this exact setup is followed can be found in \cite{liu2022convnet}. We also have downsampling blocks after each level in the ConvNext encoder. The number of ConvNext blocks is a hyperparameter. However, for simplicity we fixed the number of total levels as 4 which means the downsampling is done only 4 times throughout the encoder. The features extracted from this encoder have both spatial and temporal information and are passed to the bottleneck mixer.

\subsection{Bottleneck}

The bottleneck of both the branches consist of mixer networks with the same basic design.
%We use the same design of mixer architecture in both the branches but  they are separate instances encoding different information.
The mixer network takes in tokens $T$ as input and processes them using two different multi-layer perceptrons (MLPs). First, the input tokens are normalized and then mixed across the token dimension. The process can be summarized as:
\setlength{\belowdisplayskip}{0pt} \setlength{\belowdisplayshortskip}{0pt}
\setlength{\abovedisplayskip}{0pt} \setlength{\abovedisplayshortskip}{0pt}
\begin{equation}
    T_{TM} = MLP_{TM}(LN(T_{in})) + T_{in},
\end{equation}
where $T_{TM}$ represents the tokens extracted after Token Mixing (TM), $T_{in}$ represents the input tokens, and LN represents layer normalization \cite{ba2016layer}. Note that there is also a skip connection between the input to the mixer and the output from token mixing MLP. Token mixing encodes the relationship between individual tokens. Afterwards, the tokens are flipped along the $C$ axis and fed into another MLP to learn dependencies in the $C$ dimension \cite{tolstikhin2021mlp}. This is called channel mixing and is formulated as follows:
\begin{equation}
    T_{out} = MLP_{CM}(LN(T_{TM})) + T_{TM},
\end{equation}
where $T_{out}$ represents the output tokens and $CM$ denotes channel mixing. The MLP block comprises of two linear layers that are activated by GeLU \cite{hendrycks2016gaussian}. The initial linear layer converts the number of tokens/channels into an embedding dimension, while the second linear layer brings them back to their original dimension. The selection of the embedding dimension and the number of mixer blocks for the bottleneck is done through hyperparameter tuning.

\subsection{Recurrent training}

Recurrent setups generally refer to a type of configuration or arrangement that is repeated or ongoing \cite{sherstinsky2020fundamentals}. In real-time video enhancement, the original video stream has to be enhanced on-the-fly which means we have the information of all the enhanced frame till the current time instance.
%As there isn't much of a difference in terms of context in between nearby frames, 
The enhanced frames from the previous time step has valuable information that could be leveraged for the current prediction for increased efficiency. Leveraging previous frame prediction can also help in increasing the temporal stability of the predictions as the current predictions gets conditioned on the previous predictions. Although it is possible to use multiple previous frames in a recurrent setup, we have chosen to only use the most recent prediction for the sake of efficiency. An overview of this setup is illustrated in Fig \ref{fig:overview}.

\begin{figure}[htbp]
    \centering
\includegraphics[width=1\linewidth, page=1]{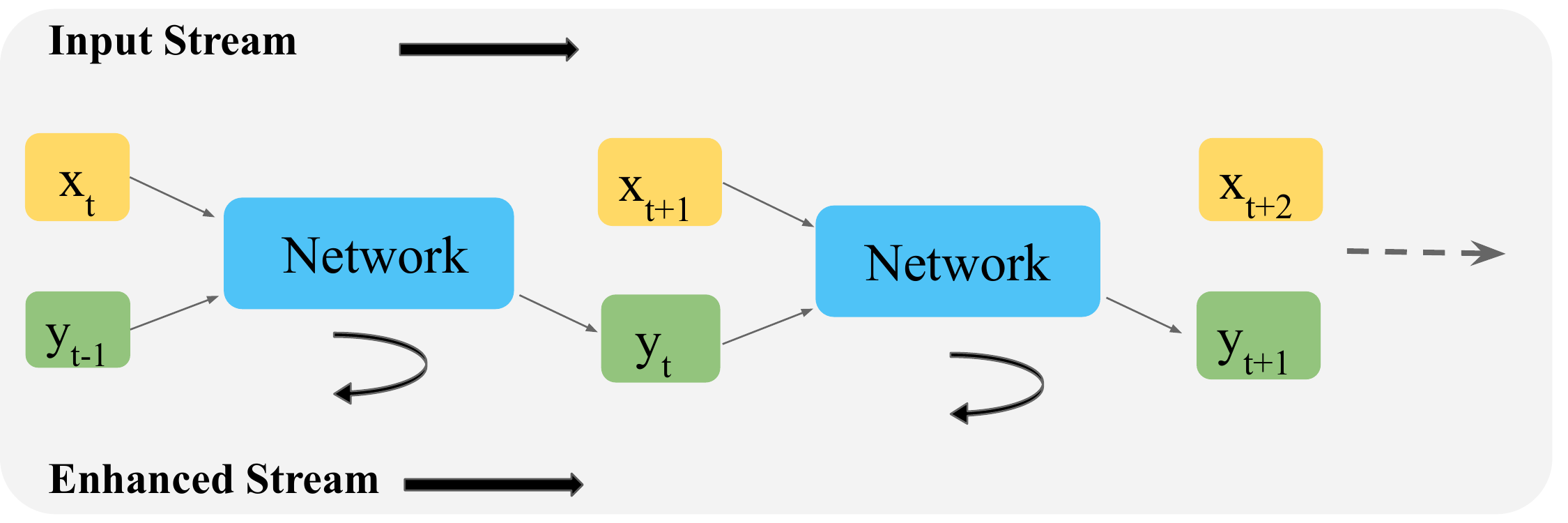}
    \vspace{-2 em}
    \caption{Overview of the proposed recurrent setup where $x_t$ is the current input frame, $y_{t-1}$ is the previous prediction, and $y_t$ represents the current prediction.}
    \label{fig:overview}
\end{figure}

%Including more than 2 frames in the input increases the number of channels in the input and so increases the complexity of the model.

In the following, we elucidate how we leverage the recurrent setup to output the enhanced frame. Let us define the original input stream as $X = \{x_0, x_1, ...., x_t\}$ where $X$ denotes the video and $x$ denotes the individual frames. The frames start from the initial frame $x_0$ to the current time frame $x_t$. Similarly, we also define the enhanced video stream represented as $Y = \{y_0, y_1, ...., y_{t-1}\}$ where $Y$ denotes the enhanced video stream and $y$ denotes the individual enhanced frames. These enhanced frames go from the initial time step $y_0$ to the previous time frame $y_{t-1}$. So, to find the enhanced prediction of the current frame $y_t$, we make use of current degraded frame $x_t$ and the previous enhanced frame $y_{t-1}$. These images are sent to the  network to output $y_t$. In the context of training, a single feed forward step involves using the input values $x_t$ and $y_{t-1}$ to make a prediction for the output value $y_t$. When processing a video, multiple feed forward steps are used in a sequential manner to predict the output values for all frames in the video. Similarly, during backpropagation, the gradients are propagated backwards through the network, starting from the last frame and moving towards the first frame of the video.
%This way, the live enhanced stream (behind by only one time-step) helps to enhance the degraded steam. 
Note that there is a corner case for the first frame while predicting $y_0$. To circumvent it, we use the ground truth frame as the initial prediction to kick-start the training. %During inference, we just duplicate the first frame for the initial frame prediction. From there on, the network gets rolled to predict all the remaining frames. %We also notice that the performance is still stable even if we utilize just the initial degraded frame without considering previous prediction for the first frame prediction. %We note that one could also warp the previous frame to the current frame to further improve performance but we do not do that due to . %As our works tries to minimize the computation as much as possible, we do not use any advanced techniques which have the potential to further improve performance at the cost of computation.

\section{Experiments and Results}

\subsection{Datasets}

We note that there exists several video super-resolution, deblurring, and denoising datasets like REDS \cite{nah2019ntire}, DVD \cite{su2017deep}, GoPro \cite{nah2017deep}, DAVIS \cite{khoreva2019video}, Set8 \cite{tassano2019dvdnet}, etc. However, these datasets focus on just one degradation at a time like deblurring or denoising. As we focus on the problem of generic video enhancement of live videos, presence of multiple degradations is very common. Also, a major use case for real-time video enhancement is video conferencing where the video actually contains the torso/face of the person. To reflect these real-world scenarios, we curate two datasets for the task of video enhancement: i) PortraitVideo and ii) FullVideo.

\noindent \textbf{PortraitVideo:} We build PortraitVideo on top of TalkingHeads \cite{wang2021one} which is a public dataset that uses Youtube videos and processes them using face detectors to obtain just the face. Here, the frame is fixed allowing only the movement of the head to be captured, which simulates a scenario where the camera is fixed during video calls\rg{Say how ours is different from that? Also, say that we use face cropping similar to FFHQ dataset?}\jmj{We follow their approach, so reworded. Also added cropping part}. The face region is then is cropped similar to face image datasets like FFHQ \cite{karras2019style}. Also, we note that TalkingHeads consists of a lot of non-human faces like cartoons and avatars as well. Further, a lot of videos are of very low quality and hence unsuitable for training or evaluation of video restoration. So, we curate PortraitVideo by skipping low quality videos and pick 113 face videos for training and 20 face videos for testing. We fix the resolution of the faces to $384 \times 384$. The videos are processed at 30 frames per second (FPS) with a total of $150$ frames per video. We use a mixture of degradations like blur with varying kernels, compression artifacts, noise, small distortions in brightness, contrast, hue, and saturation. The exact details of these degradations can be found in the supplement.
%We use a random mixture of these degradations for every video to result in diverse low quality videos to emulate real-world scenarios. The exact configurations about these degradations are picked taking into account what kinds of degradations are usually found in video calls. The exact configurations of these degradations can be found in the supplementary. 

\noindent \textbf{FullVideo:} We develop this dataset using high quality videos collected from Youtube videos. The video IDs are taken from TalkingHeads dataset however we do not use any of the pre-processing techniques from the TalkingHeads dataset so that the original information of the scene is maintained. We also manually filter to keep only high quality videos. There are $132$ training videos and $20$ testing videos, and all videos are $720 \times 1280$, 30 FPS and 128 frames long. We apply similar degradations as PortraitVideo for this dataset. The major difference is that this dataset is of a higher resolution and captures more context around the face, including the speaker's body and the rest of the scene. 
%\rg{Needs more details on how the videos were selected. Also, I am not sure if we would be able to release the data.}\jmj{added in the beginning}

\begin{table*}[]
\centering
\caption{Comparison of quantitative results of ReBotNet with previous methods. $\dagger$ represents the default configuration from paper and public code. S, M, L represent the small ($\sim$10G),  medium ($\sim$50G), and large ($\sim$400G) FLOPs regimes.   }
\resizebox{0.85\textwidth}{!}{%
\begin{tabular}{c|ccccccc}
\hline
                         &                          &                       & \multicolumn{1}{c}{}                        & \multicolumn{2}{c}{\cellcolor[HTML]{DAE8FC}PortraitVideo} & \multicolumn{2}{c}{\cellcolor[HTML]{FFCE93}FullVideo} \\  
\multirow{-2}{*}{Method} & \multirow{-2}{*}{GFLOPs ($\downarrow$) } & \multirow{-2}{*}{Latency (in ms) ($\downarrow$)} & \multicolumn{1}{c}{\multirow{-2}{*}{Param (in M) ($\downarrow$) }} & PSNR ($\uparrow$)                   & \multicolumn{1}{c}{SSIM ($\uparrow$)}          & PSNR ($\uparrow$)              & \multicolumn{1}{c}{SSIM ($\uparrow$)}             \\ \hline
FastDVDNet  $\dagger$  \cite{tassano2020fastdvdnet}            &  367.81                   &  36.23               & 1.12                                          &  28.88                  &   0.8516                           & 29.56                  &  0.8577                                    \\
VRT  $\dagger$ \cite{liang2022vrt}             & 2054.32                    & 781.15                &   19.62                                       & 31.70                  & 0.8835                             &  33.49                 & 0.9140                                      \\
BasicVSR++ $\dagger$  \cite{chan2022basicvsr++}             &   157.53                  &   49.55             & 9.3                                        &  31.26                 &  0.8739                           & 33.10                   &  0.9078                                    \\
RVRT $\dagger$   \cite{liang2022recurrent}            & 396.29                    & 52.30               & 13.57                                        & 31.92                  & 0.8870                             &  33.79                 & 0.9191                                     \\ \hline
FastDVDNet (S) \cite{tassano2020fastdvdnet}              & 15.85                    & 30.51                & \textbf{0.46}                                         & 27.97                   & 0.8384                             & 28.16                  &  0.8459                                    \\ 
VRT (S)  \cite{liang2022vrt}                   & 15.22                    & 48.73                & 2.45                                         & 30.80                   & 0.8681                             & 31.85            & 0.8901                               \\
BasicVSR++  (S)  \cite{chan2022basicvsr++}                 & 19.05                        & 29.08                     & 1.76                                            & 30.90                       & 0.8705                                  &  32.78                &  0.8950                                   \\
RVRT (S)  \cite{liang2022recurrent}                  & -                        & -                     & -                                            & -                       & -                                  & -                 & -                                     \\
\rowcolor[HTML]{EFEFEF} 
RebotNet  (S)               & \textbf{13.02}           & \textbf{13.15}        & 3.8                                          & \textbf     {31.25}          & \textbf{0.8778}                    &  \textbf{33.45}                 &  \textbf{0.9113}                                     \\ \hline
FastDVDNet  (M) \cite{tassano2020fastdvdnet}            & 64.51                    & 33.89                & \textbf{0.68}                                         & 28.52                   & 0.8405                             & 29.35                  &  0.8528                                     \\
VRT  (M)  \cite{liang2022vrt}                   & 60.18                    & 58.89                & 3.99                                         & 30.98                   & 0.8701                             &  32.35                &  0.8987                                     \\
BasicVSR++  (M) \cite{chan2022basicvsr++}                    & 60.93                    & 41.18 & 4.29                                         & 31.19                   & 0.8729                             & 33.04                &  0.9051                                     \\
RVRT  (M)   \cite{liang2022recurrent}                 & 62.42                    & 35.93                & 2.66                                         & 31.60                   & 0.8821                             & \textbf{33.59}                  &  0.9145                                     \\
\rowcolor[HTML]{EFEFEF} 
RebotNet (M)                 & \textbf{56.06}           & \textbf{15.02}        & 6.86                                         & \textbf{31.85}          & \textbf{0.8865}                    &  33.45                 &   \textbf{0.9168}                                    \\ \hline
FastDVDNet (L)  \cite{tassano2020fastdvdnet}             & 416.90                   & 37.14                 & \textbf{1.19}                                         & 28.93                   & 0.8537                             &  29.68                &  0.8593                                    \\
VRT  (L)  \cite{liang2022vrt}                   & 419.32                   & 91.74                & 20.96                                        & 31.09                   & 0.8729                             & 32.68                  &        0.9014                              \\
BasicVSR++ (L) \cite{chan2022basicvsr++}                     & 403.22                  & 73.32                & 24.55                                        &  31.40                  &  0.8775                          &  33.31                & 0.9126                                   \\
RVRT  (L)  \cite{liang2022recurrent}                  & 396.29                   & 52.30                & 13.57                                       & 31.92                   & 0.8870                             & \textbf{33.79}                   & 0.9191                                      \\
\rowcolor[HTML]{EFEFEF} 
RebotNet  (L)               & \textbf{363.76}          & \textbf{19.98}        & 41.3                                         & \textbf{32.13}          & \textbf{0.8902}                    & 33.65                   & \textbf{0.9199}                                       \\ \hline
\end{tabular}
}
\label{res}
\end{table*}

\subsection{Implementation Details}

We prototype our method using PyTorch on NVIDIA A100 GPU cluster. ReBotNet is trained with a learning rate of $4e^{-4}$ using Adam optimizer, and a cosine annealing learning rate scheduler with a minimum learning rate of $1e^{-7}$. The training is parallelized across 8 NVIDIA A100 GPUs, with each GPU processing a single video. The model is trained for 500,000 iterations.  For fair comparison with existing methods, we only use the commonly used Charbonnier loss \cite{barron2019general} to train all models. More configuration details of the architecture can be found in the supplementary.

\subsection{Comparison with previous works}

We compare ReBotNet against multiple recent methods. Recurrent Video Restoration Transformer (RVRT) \cite{liang2022recurrent} is the current SOTA method across many tasks like deblurring, denoising, super-resolution, and video-frame interpolation. We also compare against Video Restoration Transformer (VRT) \cite{liang2022vrt}, the SOTA convolution-based video super-resolution method BasicVSR++ \cite{chan2022investigating}, and the fastest deblurring method FastDVD for fair comparison. We retrain all these methods on the new datasets PortraitVideo and FullVideo using their publicly available code.

\begin{table}[htbp]
\caption{Comparison of ReBotNet with previous methods on public datasets. Numbers correspond to PSNR / SSIM.}
\centering
\resizebox{0.3\textwidth}{!}{%
\begin{tabular}{c|c|c}
\hline
Method     & DVD \cite{su2017deep}            & GoPro \cite{nah2017deep}         \\ \hline
DeepDeblur \cite{nah2017deep} & 29.85 / 0.8800 & 38.23 / 0.9162 \\
EDVR \cite{wang2019edvr}       & 31.82 / 0.9160 & 31.54 / 0.9260 \\
TSP \cite{pan2020cascaded}       & 32.13 / 0.9268 & 31.67 / 0.9279 \\
PVDNet \cite{son2021recurrent}    & 32.31 / 0.9260 & 31.98 / 0.9280 \\
VRT  \cite{liang2022vrt}      & 34.24 / 0.9651 & 34.81 / 0.9724 \\
RVRT \cite{liang2022recurrent}      & 34.30 / 0.9655 & 34.92 / 0.9738 \\
ReBotNet   & 34.28/ 0.9656  & 34.90 / 0.9734
\end{tabular}
}
\label{pub}
\end{table}

Initially,  we conducted experiments on the new datasets PortraitVideo and FullVideo using the default configurations of VRT, RVRT, BasicVSR++, and FastDVD, as provided in their publicly available code\rg{Do we report these numbers somewhere? We should have them handy, at least for the rebuttal}\jmj{yes, they are in first few rows of Table 1} as seen in the first few rows of Table \ref{res}. It is important to mention that these models have different levels of floating-point operations (FLOPs). Therefore, to ensure a fair comparison, we assessed the performance of ReBotNet in different FLOP regimes in comparison to the previous methods. This approach helped us gain a comprehensive understanding of the performance of these models across different levels of FLOPs. We pick the embedding dimension across different levels of the network as the hyper-parameter to change the FLOPs \cite{kondratyuk2021movinets}. We acquire different configurations of FLOPs regimes of Small (10Gs), Medium (50Gs), and Large (400Gs). The exact configuration details can be found in the supplementary material. Note that RVRT does not have a S configuration as it is infeasible to scale down the model near 10 GFLOPs due to its inherent design. It should also be noted that for each configuration, we ensured that the computational complexity of ReBotNet remained lower than that of the other models being compared. To provide an example, when evaluating models in the medium regime, we compared ReBotNet, which had a complexity of 56.06, with VRT, which had a complexity of 60.18, and RVRT, which had a complexity of 62.42. In all of our experiments, we used a consistent number of frames, which was set to 2 for all models except for FastDVD, which was designed to process 5 frames. To evaluate the models, we compute the PSNR and SSIM for each individual frame of a video, and then average these values across all frames within the video. We then calculated the mean PSNR and SSIM across all videos and present these results in Table \ref{res}. We use the high-quality frame as the first frame for all these methods while performing the inference. Additionally, we measure the inference time for each method by forwarding 2 frames of dimensions $(384,384)$ through the network. To obtain the latency, we perform GPU warm-up for 10 iterations and then feed-forward the clip 1000 times, reporting the average. The latency was recorded on a NVIDIA A100 GPU. We also report the number of parameters for each model.

\begin{figure*}
    \centering
    \includegraphics[width=0.95\linewidth, page=1]{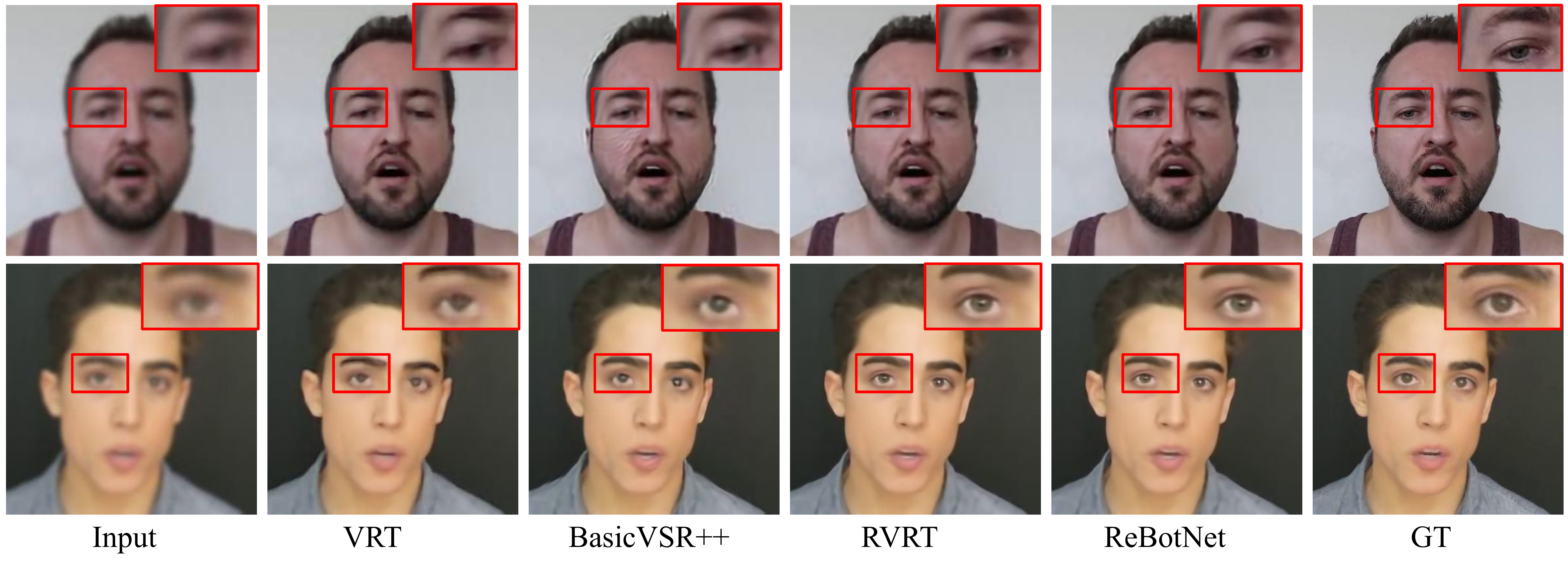}
    \vspace{-1 em}
    \caption{Qualitative Results on \textit{PortraitVideo} dataset. Please zoom in for better visualization.}
    \label{fig:th}
    \vspace{-0.5 em}
\end{figure*}

\begin{figure*}
    \centering
    \includegraphics[width=0.95\linewidth, page=1]{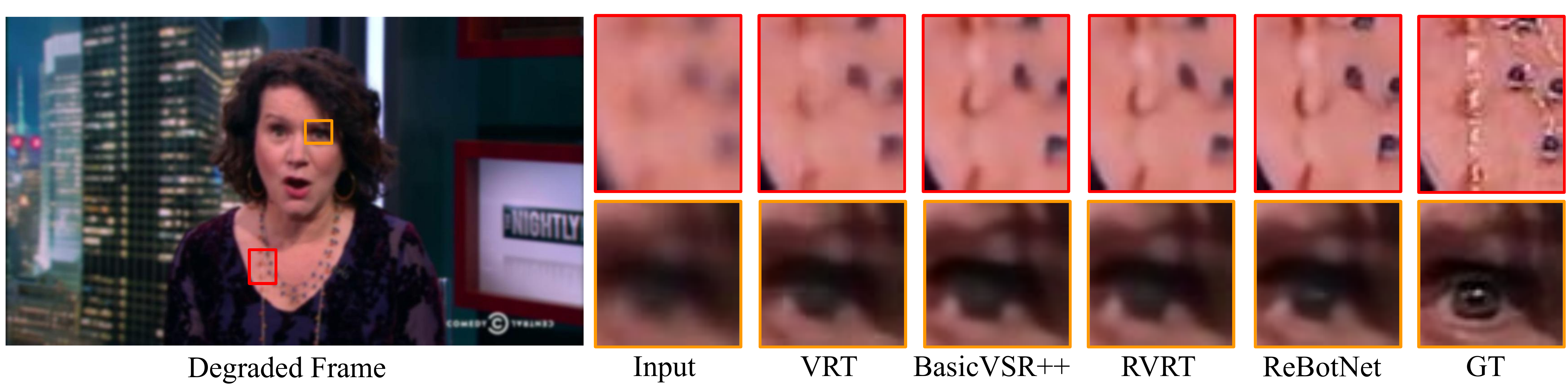}
    \vspace{-1 em}
    \caption{Qualitative Results on \textit{FullVideo} dataset. Please zoom in for better visualization.}
    \label{fig:viden}
    \vspace{-1 em}
\end{figure*}

Table \ref{res} demonstrates that our method outperforms most previous approaches in terms of PSNR and SSIM, while using less computational resources across most regimes for both datasets. A significant advantage of our model is its fast inference speed, which is 2.5x faster than the previous best performing method, RVRT. We also note that we get better results than the original implementations which have way more computations (as seen in first few rows of Table \ref{res}) \rg{Is it captured in the table?}\jmj{yes, pointed to the exact location}. The efficiency of ReBotNet comes because of its effective design while also employing token mixing mechanisms by using mixers. The main contribution towards computation 
 in transformer-based methods like RVRT and VRT come from the self-attention mechanism acting at original scale of the image. Note that we do not use self-attention but replace it with a careful design choice that matches (or even exceeds) its performance. We also conduct experiments on single degradation public  datasets like DVD, GoPro and report the results in Table \ref{pub} \rg{Should we add latency numbers to this table as an additional column?}\jmj{It would be great to show but I am having difficulty finding codes for some of them, so I have just added a line pointing to table 1 for latency. Anyways, only VRT and RVRT compare close to ours and we are clearly faster then them as seen in Table 1.}. For this, we use ReBotNet (L) and compare against the default configurations of previous methods. It can be observed that we obtain a competitive performance in spite of low latency of our model, which can be already seen in Table \ref{res}. 

In Figures \ref{fig:th} and \ref{fig:viden}, we present qualitative results from PortraitVideo and FullVideo dataset. It can be observed that our method is better than previous methods in terms of quality. The enhanced details are much visible in ReBotnet when compared to other methods. The results are taken from the medium configurations of each model. More results can be found in the supplementary material. \rg{Point to more results in the supplementary?}\jmj{added}

% \begin{figure*}
%     \centering
%     \includegraphics[width=1\linewidth, page=1]{/figs/viva_res.pdf}
%     \caption{Results on VIVA HQ dataset}
%     \label{fig:viva}
% \end{figure*}

\subsection{User Study}
To validate the perceptual superiority of ReBotNet for video enhancement, we conducted a user study on the M configuration models on PortraitVideo dataset. We compare our approach to each competing method in a one-to-one comparison. We recruited 3 experts with technical experience in conference video streaming and image enhancement. Each expert evaluated on average 80 video comparisons across four baseline methods. For each comparison, we showed output videos of our method and one competing method, played both videos simultaneously and asked the user to rate which video had a higher quality with the corresponding scores ("much worse", -2), ("worse", 1), ("same", 0), ("better", 1) and ("much better", 2). We calculated the mean score and 95\% confidence intervals for paired samples and report them in Table \ref{user}. The user study demonstrates the superiorty of our method. Despite RVRT being the closest second, our method is still preferred over it while also being more efficient and faster.

\begin{table}[]
\centering
\caption{User study results on PortraitVideo dataset.}
\vspace{-1 em}
\resizebox{0.4\textwidth}{!}{%
\begin{tabular}{c|cc}
\hline
Method     & Preference for ReBotNet & $95 \%$ Confidence Interval\\ \hline
FastDVDNet &   $+$           1.83   &  0.059                            \\
VRT        &  $+$         1.61        &  0.088                        \\
BasicVSR++ &   $+$   1.63              &  0.105                      \\
RVRT       &   $+$ 0.08                 &  0.073                     
\end{tabular}
}
\label{user}
\vspace{-2 em}
\end{table}

\section{Discussion}

\begin{table*}[h]
\caption{Analysis on the (a) number of embedding dimension in Mixer (b) depth of the botteneck (c) number of frames taken.}
\vspace{-1 em}
		 \begin{minipage}{0.333\linewidth}
		 \centering
		 \vspace{0em}
  
   \resizebox{1\linewidth}{!}{%
\begin{tabular}{c|cccc}
\hline
Embedding & PSNR ($\uparrow$) & SSIM ($\uparrow$)& GFLOPs ($\downarrow$) & Latency ($\downarrow$)\\ \hline
128       & 31.79     & 0.8851     & 55.50       & 14.85        \\
\rowcolor[HTML]{EFEFEF} 256       & 31.85     & 0.8865     &  56.06      &  15.02       \\
512       & 31.90     & 0.8869     &  56.60      & 15.27        \\
728       & 31.89     & 0.8869     & 57.14       &  15.36      
\end{tabular}
}
 \caption*{ (a) }
\label{comp}
	\end{minipage}%
 \begin{minipage}{0.333\linewidth}
		 \centering
		 \vspace{0em}
   \resizebox{0.91\linewidth}{!}{%
   \begin{tabular}{c|cccc}
\hline
Depth & PSNR ($\uparrow$)& SSIM ($\uparrow$)& GFLOPs ($\downarrow$)& Latency ($\downarrow$)\\ \hline
2     & 31.83     & 0.8864     &  55.50      & 14.67        \\
\rowcolor[HTML]{EFEFEF} 4     & 31.85     & 0.8865     &  56.06      &  15.02       \\
6     & 31.87     & 0.8866     &  57.14      &  15.31       \\
8     & 31.81     & 0.8861     &  58.08      &  16.34      
\end{tabular}
}
 \caption*{ (b) }
   \label{comp2}
	\end{minipage}%
 \begin{minipage}{0.333\linewidth}
		 \centering
		 \vspace{0em}
   \resizebox{0.91\linewidth}{!}{%
 \begin{tabular}{c|cccc}
\hline
Frames & PSNR ($\uparrow$)& SSIM ($\uparrow$)& GFLOPs ($\downarrow$)& Latency ($\downarrow$)\\ \hline
1      & 29.56     & 0.8586     & 55.50       & 14.85        \\
\rowcolor[HTML]{EFEFEF} 2      &  31.85    & 0.8865 & 56.06       &  15.02       \\
3      & 31.88     & 0.8871     &  57.14      &  15.16       \\
4      & 31.92     &  0.8874    &  58.08      &  15.40      
\end{tabular}
}
 \caption*{ (c) }
\label{comp3}
	\end{minipage}%
 \label{hype}
 \vspace{-2.5 em}
 \end{table*}
\noindent \textbf{Analysis on ReBotNet:} In order to elucidate our design decisions for ReBotNet, we carry out a set of experiments using various parameter configurations, which affect both the performance and computational aspects of the model. These experiments are conducted on the PortraitVideo dataset, using ReBotNet (M) as the base configuration. Table \ref{hype} illustrates these results where gray rows correspond to the configuration of the actual implementation of ReBotNet (M). We analyze the performance along with computation and latency on different configurations of embedding dimension in Mixer (Table \ref{hype}.a), depth of the bottleneck (Table \ref{hype}.b), and the number of frames (Table \ref{hype}.c).

\noindent \textbf{Ablation Study:} In order to investigate the contribution of each component proposed in the work, we conduct an ablation study using the PortraitVideo dataset. The results of these experiments are shown in Table \ref{abl}. First, we  use the Tubelet tokens extracted from spatio-temporal branch where we use ConvNext encoder directly with a decoder to get the prediction. Then, we consider a configuration where we use image tokens extracted using linear layers from the spatial branch directly forwarded to decoder to get the prediction. This configuration obtains the best latency however suffers from a significant drop in performance. Next, we fuse features extracted from both these branches and use the common decoder. This shows a relative improvement in terms of performance without much addition in computation. Note that here the FLOPs of fused configuration is not direct addition between FLOPs of tubulet tokens and image tokens as the decoder's computation was common in both the previous setups. Next, we add the bottleneck mixers which obtains an improvement in performance with little increase in compute. Finally, we add the recurrent training setup which adds no increase in compute but improves the performance. Our findings indicate that each individual component in ReBotNet plays a vital role.

\begin{table}[htb]
\centering
\caption{Ablation study on PortraitVideo dataset.}
\vspace{-1 em}
\resizebox{0.5\textwidth}{!}{%
\begin{tabular}{c|c|c|c|cccc}
\hline
Tub. Tok.             & Img Tok.              & Bot. Mix.     & Rec. Setup       & PSNR ($\uparrow$)                 & SSIM ($\uparrow$) & GFLOPs ($\downarrow$) & Latency ($\downarrow$) \\ \hline
                $\checkmark$       &  $\times$                     &  $\times$                       &  $\times$                       & 31.24                     &  0.8768    & 54.94       & 14.27        \\
                    $\times$    &    $\checkmark$                    & $\times$                        &   $\times$                      &  28.01                    & 0.8295     &  41.94      & 5.63        \\
                     $\checkmark$   &  $\checkmark$                      &   $\times$                      &   $\times$                      &  31.41                    & 0.8792     &  55.50      & 14.67        \\
                     $\checkmark$  & $\checkmark$                       &  $\checkmark$                      &   $\times$                      &  31.59                    & 0.8822     & 56.06       &   15.02      \\
$\checkmark$  &$\checkmark$   &$\checkmark$   &$\checkmark$   & 31.85  & 0.8865     &  56.06      &  15.02      
\end{tabular}
}
\label{abl}
\end{table}

% \subsection{Transformers vs Mixers}

% \begin{table}[htbp]
% 	\centering
% 		\caption{Comparison of Mixers with Transformer for Video Enhancement}
% 		\resizebox{1\linewidth}{!}{%
% 	\begin{tabular}{c|cc}
% 		\Xhline{2\arrayrulewidth}

% 		Method           & PSNR ($\uparrow$)   & SSIM  ($\uparrow$)                       \\ \Xhline{2\arrayrulewidth}

% 		Recurrent Bottleneck Mixer &     &  \\
% 		Recurrent Bottleneck Transformer &    &  \\
% 		\Xhline{2\arrayrulewidth}

% 	\end{tabular}
% 	}

% 	\label{den}

% \end{table}

\noindent \textbf{FPS and Peak Memory Usage:} In Figure \ref{fig:chart1}, we provide a comparison of ReBotNet's frames per second (FPS) rate and peak memory usage with previous methods.  For this analysis, we consider feed forward of 2 frames of resolution $384 \times 384$ and consider RebotNet (L) configuration with original implementations for the previous methods. It can be observed that our method has a FPS that is real-time while also not occupying much memory. We note that 30 FPS is considered real-time for applications like video conferencing. Also, ReBotNet has one of the least memory requirements compared to other methods due to its efficient design and implementation.

\begin{figure}
    \centering
    \includegraphics[width=0.85\linewidth]{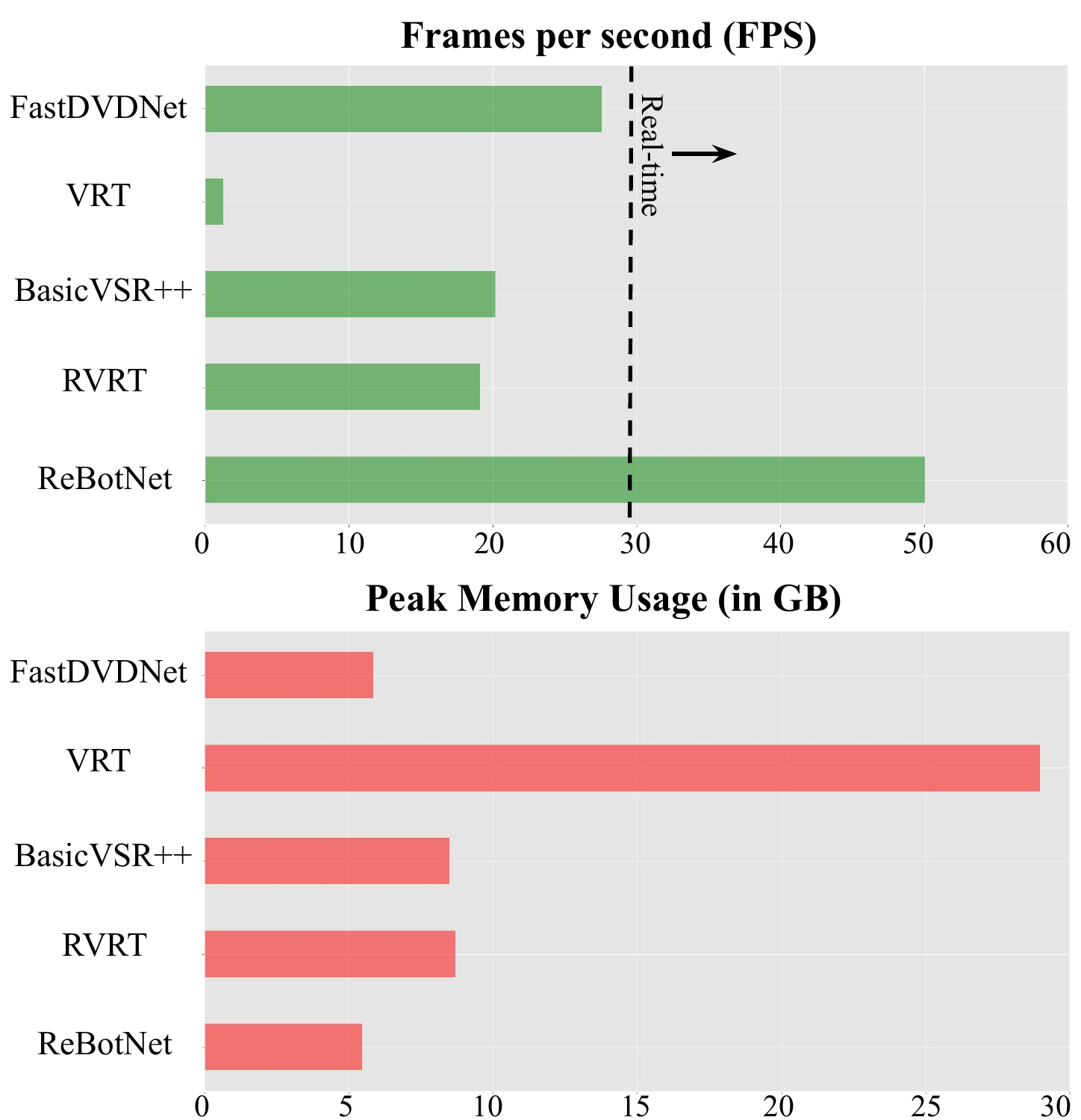}
    \vspace{-1 em}
    \caption{Comparison chart of ReBotNet (L) against default configurations of previous methods for Frames Per Second (FPS) and Peak Memory Usage (in GB), as measured on NVIDIA A100 GPU for $ 2 \times 3 \times 384 \times 384$ resolution.}
    \label{fig:chart1}
\vspace{-0.6 em}
\end{figure}

\noindent \textbf{Limitations:} Our method  is not ideal in terms of the number of parameters. This is not a concern for applications such as video calls or live streams, where processing is usually performed in the cloud. However, if the method is to be used for edge applications, it is necessary to optimize the number of parameters. %The number of parameters of ReBotNet is 6.86 M compared to VRT's 3.99 M and RVRT's 2.66 M all at around 60 GFLOPs. In terms of qualitative results, our method produces a better quality for highly degraded videos. However, the same is not true for light degradations. In those cases, the performance difference between the state-of-the-art methods and the proposed method is negligible. 
To focus on improvement due to the proposed architecture alone, we only used Charbonnier loss to train all models. Additional losses like the perceptual loss \cite{johnson2016perceptual} \rg{Add citation}\jmj{added} can be applied to further improve the results. 

\section{Conclusion}

In this paper, we proposed a novel approach for real-time video enhancement by proposing a new framework: Recurrent bottleneck mixer network (ReBotNet). ReBotNet combines the advantages of both recurrent setup and bottleneck models, allowing it to effectively capture temporal dependencies in the video while reducing the computational complexity and memory requirements. We evaluated the performance of ReBotNet on multiple video enhancement datasets. The results showed that our proposed method outperformed state-of-the-art methods in terms  computational efficiency while matching or outperforming them in terms of visual quality. %Further research could explore the potential applications of the proposed network for other video processing tasks. 

%%%%%%%%% REFERENCES
{\small
\bibliographystyle{ieee_fullname}
\bibliography{egbib}
}

\newpage

\appendix

\section{Configurations of ReBotNet}

In the main paper, we mentioned we conducted experiments with different FLOPs regimes for all the methods. We did that by controlling the depth of the bottleneck and the embedding dimension of different methods to get the required FLOPs. In Tables \ref{1}, \ref{2}, and \ref{3} we provide the exact configurations of ReBotNet - S,M, and L respectively. More analysis on the dependence of these parameters were provided in the main paper.

\begin{table}[htbp]
\centering
\caption{Configuration of ReBotNet-S.}
\begin{tabular}{c|cc}
\hline
Block                       & Type                 & Value             \\ \hline
\multirow{3}{*}{Branch I}   & Number of Layers     & 4                 \\
                            & Depths per layer     & {[}4,4,4,4{]}     \\
                            & Embedding dimensions & {[}28,36,48,64{]} \\
\multirow{2}{*}{Branch II}  & Patch size           & 1                 \\
                            & Embedding Dimension  & 256               \\
\multirow{3}{*}{Bottleneck} & Depth                & 4                 \\
                            & Input Dimension      & 64                \\
                            & Hidden Dimension     & 728              
\end{tabular}
\label{1}
\end{table}

\begin{table}[htbp]
\centering
\caption{Configuration of ReBotNet-M.}
\begin{tabular}{c|cc}
\hline
Block                       & Type                 & Value             \\ \hline
\multirow{3}{*}{Branch I}   & Number of Layers     & 4                 \\
                            & Depths per layer     & {[}4,4,4,4{]}     \\
                            & Embedding dimensions & {[}64,80,108,116{]} \\
\multirow{2}{*}{Branch II}  & Patch size           & 1                 \\
                            & Embedding Dimension  & 256               \\
\multirow{3}{*}{Bottleneck} & Depth                & 4                 \\
                            & Input Dimension      & 116                \\
                            & Hidden Dimension     & 728              
\end{tabular}
\label{2}
\end{table}

\begin{table}[htbp]
\centering
\caption{Configuration of ReBotNet-L.}
\begin{tabular}{c|cc}
\hline
Block                       & Type                 & Value             \\ \hline
\multirow{3}{*}{Branch I}   & Number of Layers     & 4                 \\
                            & Depths per layer     & {[}5,5,5,4{]}     \\
                            & Embedding dimensions & {[}172,180,188,196{]} \\
\multirow{2}{*}{Branch II}  & Patch size           & 1                 \\
                            & Embedding Dimension  & 256               \\
\multirow{3}{*}{Bottleneck} & Depth                & 4                 \\
                            & Input Dimension      & 64                \\
                            & Hidden Dimension     & 728              
\end{tabular}
\label{3}
\end{table}

\section{Configuration of Baselines}

We used the publicly available codes for the original implementations of FastDVDNet, BasicVSR++, VRT, and RVRT; the results of which can be seen in Table 1 of the main paper. For the S,M and L configurations we use the same configurations of the original implementations but change the embedding dimensions. These changes have been illustrated in Tables \ref{4}, \ref{5}, and \ref{6}. OG means the original implementation. Note that RVRT does not have a S configuration as even with embedding dimensions of $[1,1,1]$, the FLOPs does not hit the range of 10 GFLOPs.
\begin{table}[]
\centering
\caption{Configurations of VRT.}
\resizebox{1\columnwidth}{!}{%
\begin{tabular}{c|c}
\hline
Method  & Embedding Dimension                             \\ \hline
VRT - S & {[}24,24,24,24,24,24,24,24,24,24{]}             \\
VRT - M & {[}48,48,48,48,48,48,48,48,48,48{]}             \\
VRT - L & {[}180,180,180,180,180,180,120,120,120,120{]}
\\
VRT - OG & {[}180,180,180,180,180,180,120,120,120,120,120,120,120{]}
\end{tabular}
}
\label{4}
\end{table}
\begin{table}[]
\centering
\caption{Configurations of RVRT.}
\begin{tabular}{c|c}
\hline
Method   & Embedding Dimension \\ \hline
RVRT - S & -                   \\
RVRT - M & {[}36,36,36{]}      \\
RVRT - L & {[}192,192,192{]}  \\
RVRT - OG & {[}192,192,192{]} 
\end{tabular}
\label{5}
\end{table}
\begin{table}[]
\centering
\caption{Configurations of FastDVDNet.}
\begin{tabular}{c|c}
\hline
Method   & Embedding Dimension \\ \hline
FastDVDNet - S & {[}32, 48, 72, 96{]}                   \\
FastDVDNet - M & {[}64, 80, 108, 116{]}      \\
FastDVDNet - L & {[}96, 112, 132, 144{]}  \\
FastDVDNet - OG & {[}80, 96, 132, 144{]}  
\end{tabular}
\label{6}
\end{table}
% \begin{table}[]
% \centering
% \caption{Configurations of BasicVSR++.}
% \begin{tabular}{c|c}
% \hline
% Method   & Embedding Dimension \\ \hline
% BasicVSR++ - S & {[}32, 48, 72, 96{]}                   \\
% BasicVSR++ - M & {[}64, 80, 108, 116{]}      \\
% BasicVSR++ - L & {[}96, 112, 132, 144{]}  \\
% BasicVSR++ - OG & {[}96, 112, 132, 144{]}
% \end{tabular}
% \end{table}
\begin{figure*}[htbp]
    \centering
    \includegraphics[width=0.95\linewidth, page=1]{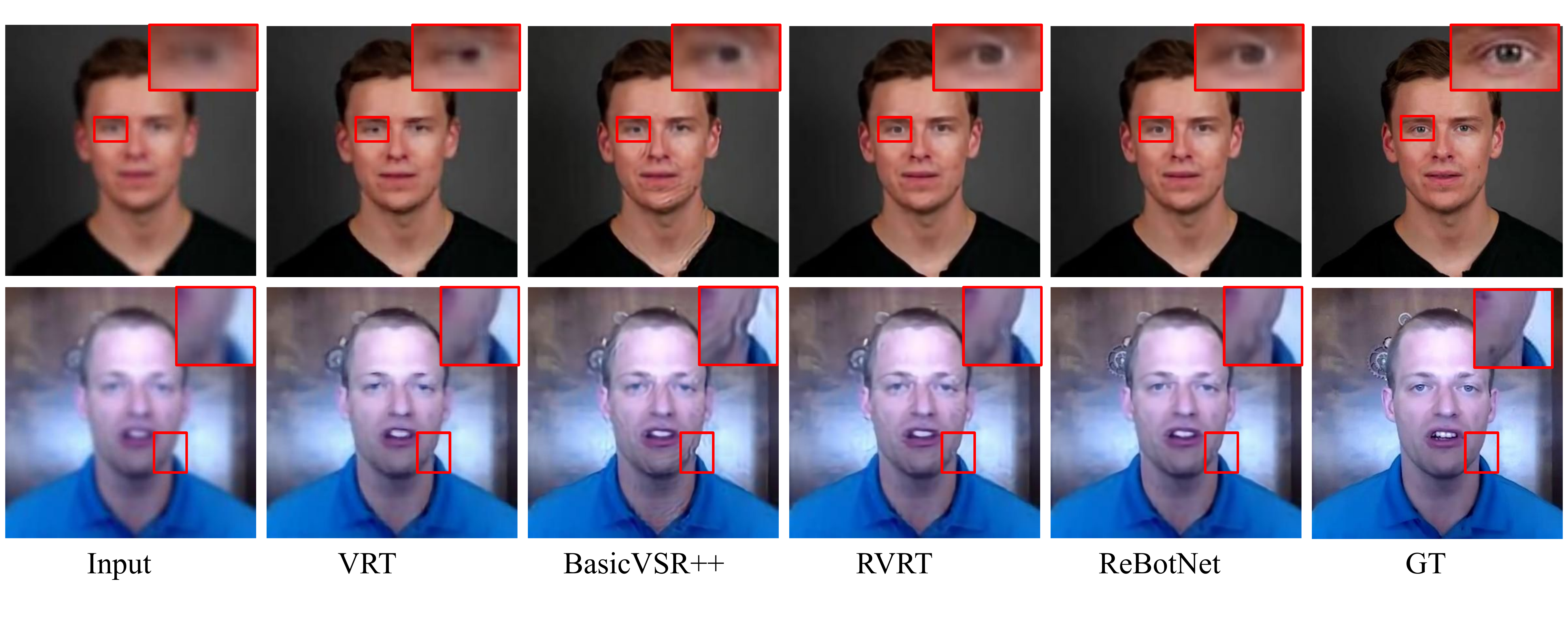}
    \vspace{-1 em}
    \caption{Qualitative Results on \textit{PortraitVideo} dataset. Please zoom in for better visualization.}
    \label{portvid}
\end{figure*}
\begin{figure*}[htbp]
    \centering
    \includegraphics[width=0.95\linewidth, page=2]{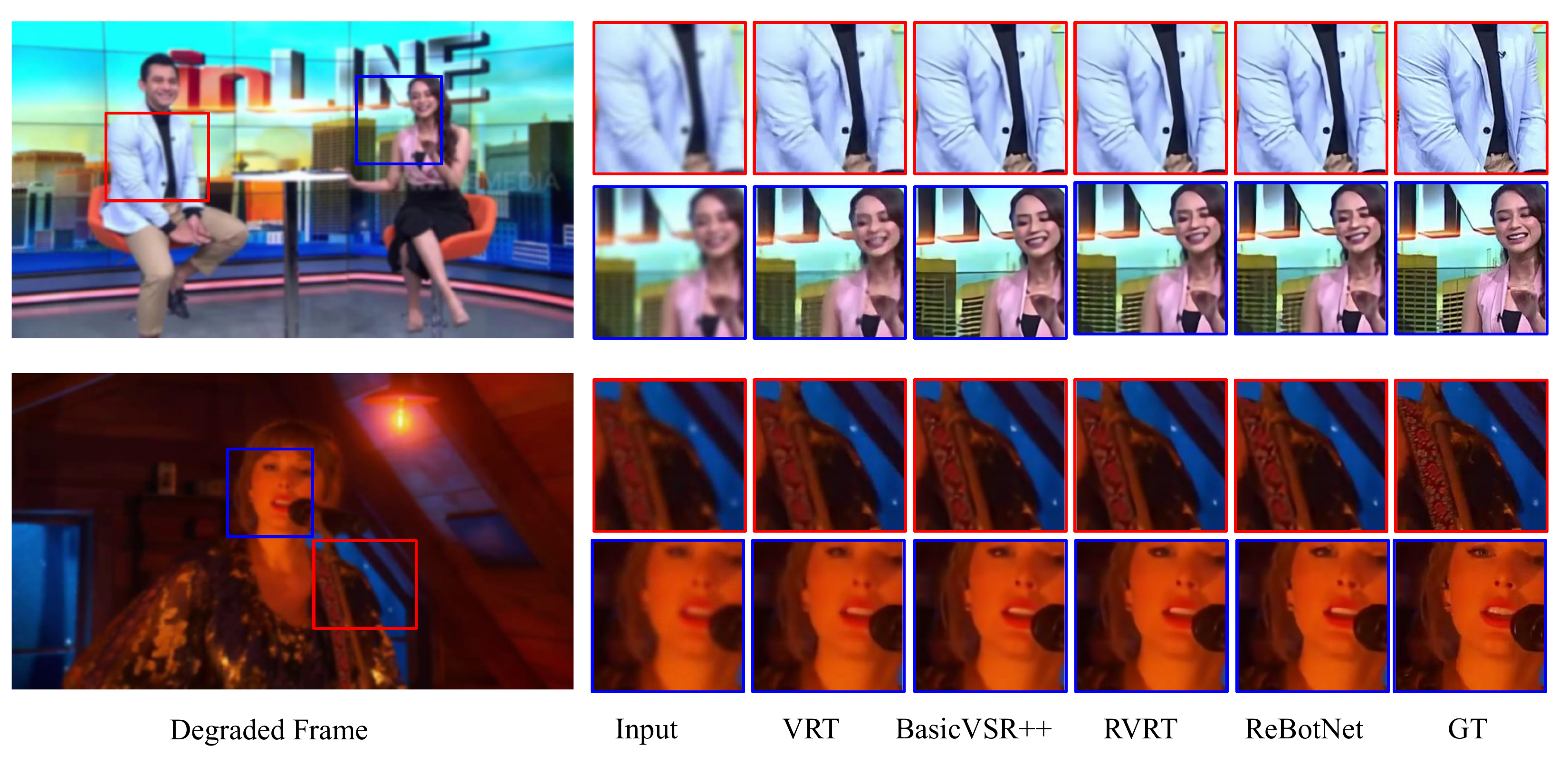}
    \vspace{-1 em}
    \caption{Qualitative Results on \textit{FullVideo} dataset. Please zoom in for better visualization.}
    \label{fullvid}
\end{figure*}

\section{Experiments on Pure Mixers}

 We observed that MLP-Mixers tend to exhibit a noticeable decline in quality when applied directly for video enhancement compared to transformer-based approaches. Using Mixers directly on large size images still takes a lot of compute and makes it difficult to achieve real-time speed.  In Table \ref{mix}, we conduct an experiment where we take VRT as the base network and convert all the transformer blocks in it to MLP-Mixers. The experiment is conducted on the DVD dataset. It can be observed that the although the computation reduces, the performance also drops significantly. And still the computation is far away from obtaining a real-time FPS. This motivates us to work towards our design of ReBotNet as seen in the main paper.

\begin{table}[]
\centering
\caption{Experiment on pure mixers.}
\begin{tabular}{c|cccc}
\hline
Method       & PSNR  & SSIM   & GFLOPs  & FPS\\ \hline
VRT          & 34.24 & 0.9651 & 2054.32  &1\\
VRT (Mixers) & 32.14 & 0.9429 & 1495.06 &2
\end{tabular}
\label{mix}
\end{table}
 \section{More Qualitative results}

In Figures \ref{portvid} and \ref{fullvid}, we provide more qualitative results on  PortraitVideo and FullVideo datasets respectively.

\section{Degradations}

In Table \ref{deg}, we provide the detailed configurations of degradations that we use in PortraitVideo and FullVideo dataset. In all the rows where there is a range, we choose a random value in the range. To get the final degradation of a sample image at hand, we choose a random combination of the degradations from Table  \ref{deg}. These values were decided to emulate degradations possible in real-world and after consulting experts working in the field of video conferencing.

\begin{table}[htbp]
\centering
\caption{Degradations used in PortraitVideo and FullVideo datasets.}
\begin{tabular}{c|c}
\hline
Type of Degradation          & Value            \\ \hline
Eye Enlarge ratio            & 1.4              \\
Blur kernel size             & 15               \\
Kernel Isotropic Probability & 0.5              \\
Blur Sigma                   & {[}0.1,3{]}      \\
Downsampling range           & {[}0.8,2.5{]}    \\
Noise amplitude              & {[}0,0.1{]}      \\
Compression Quality          & {[}70,100{]}     \\
Brightness                   & {[}0.8,1.1{]}    \\
Contrast                     & {[}0.8,1.1{]}    \\
Saturation                   & {[}0.8,1.1{]}    \\
Hue                          & {[}-0.05,0.05{]}
\end{tabular}
\label{deg}
\end{table}

\section{Reasons behind design choices in Branch I}

 We pick ConvNext blocks over basic ConvNet blocks as it has been shown that they are both efficient and effective than ConvNets. Each ConvNext block first consists of a depth-wise convolution block with kernel size of $7 \times 7$, stride $1$ and padding $3$. Using a large kernel size is to have a larger receptive field similar to non-local attention. It was observed in \cite{liu2022convnet} that
the benefit of larger kernel sizes reaches a saturation point at at $7 \times 7$. It is followed by a layer normalization and a point-wise convolution function. The point-wise convolution is basically a convolution layer with kernel size $1 \times 1$. The output of this is activated using GeLU activation and then forwarded to another point-wise convolution to get the output. More details of this why this exact setup is followed can be found in \cite{liu2022convnet}. We also have downsampling blocks after each level in the ConvNext encoder. The number of ConvNext blocks is a hyperparameter. However, for simplicity we fixed the number of total levels as 4 which means the downsampling is done only 4 times throughout the encoder. 

\end{document}